\documentclass[pdflatex,sn-mathphys]{sn-jnl}% Math and Physical Sciences Reference Style

\usepackage{amsmath,amssymb,amsfonts}
\usepackage{commath}
\usepackage{lipsum}
\usepackage{multicol}
\usepackage{pdfpages}
\usepackage{epstopdf}
\usepackage{multirow}
\usepackage{booktabs}
\usepackage{afterpage}
\usepackage{soul}
\usepackage{pifont}
\newcommand{\cmark}{\ding{51}}
\newcommand{\xmark}{\ding{55}}
\usepackage{hyperref}
\usepackage{color,soul}
\usepackage{xcolor}
\hypersetup{
     colorlinks = true,
     linkcolor = blue,
     anchorcolor = blue,
     citecolor = green,
     filecolor = magenta,
     urlcolor = cyan
     }

\jyear{2021}%

\theoremstyle{thmstyleone}%
\theoremstyle{thmstyletwo}%

\theoremstyle{thmstylethree}%

\begin{document}

\title[Multi-Modal Fusion for Sensorimotor Coordination in Steering Angle Prediction]{Multi-Modal Fusion for Sensorimotor Coordination in Steering Angle Prediction}


\author[1]{\fnm{Farzeen} \sur{Munir}}\email{farzeen.munir@gist.ac.kr}

\author[1]{\fnm{Shoaib} \sur{Azam}}\email{shoaibazam@gm.gist.ac.kr}

\author[1]{\fnm{Byung-Geun} \sur{Lee}}\email{bglee@gist.ac.kr}

\author*[1]{\fnm{Moongu} \sur{Jeon}}\email{mgjeon@gist.ac.kr}

\affil*[1]{\orgdiv{School of Electrical Engineering and Computer Science}, \orgname{Gwangju Institute of Science and Technology}, \orgaddress{\street{123 Cheomdangwagi-ro, Buk-gu},  \postcode{61005}, \state{Gwangju}, \country{South Korea}}}


\abstract{Imitation learning is employed to learn sensorimotor coordination for steering angle prediction in an end-to-end fashion requires expert demonstrations. These expert demonstrations are paired with environmental perception and vehicle control data. The conventional frame-based RGB camera is the most common exteroceptive sensor modality used to acquire the environmental perception data. The frame-based RGB camera has produced promising results when used as a single modality in learning end-to-end lateral control. However, the conventional frame-based RGB camera has limited operability in illumination variation conditions and is affected by the motion blur. The event camera provides complementary information to the frame-based RGB camera. This work explores the fusion of frame-based RGB and event data for learning end-to-end lateral control by predicting steering angle. In addition, how the representation from event data fuse with frame-based RGB data helps to predict the lateral control robustly for the autonomous vehicle. To this end, we propose DRFuser,  a novel convolutional encoder-decoder architecture for learning end-to-end lateral control. The encoder module is branched between the frame-based RGB data and event data along with the self-attention layers. Moreover, this study has also contributed to our own collected dataset comprised of event, frame-based RGB, and vehicle control data. The efficacy of the proposed method is experimentally evaluated on our collected dataset, Davis Driving dataset (DDD), and  Carla Eventscape dataset. The experimental results illustrate that the proposed method DRFuser outperforms the state-of-the-art in terms of root-mean-square error (RMSE) and mean absolute error (MAE) used as an evaluation metrics.}




\maketitle

\section{Introduction}\label{sec1}

Autonomous driving has gained tremendous attention both in vision and robotics communities over the past decade. The development of autonomous vehicles involves modular design incorporating localization, perception, planning and control \cite{azam2020system}. Since the vision modality mimics human perception more, a surrogate architecture to classical design is to learn a mapping function between the vision modality and control actuators. The learning of this mapping function broadly involves two approaches that are i) computer vision approaches and ii) imitation learning-based approach. The former approach of predicting the steering angle involves road boundaries or lane marking detection as the feature extraction mechanism. So, firstly, the images are pre-processed and followed by either road boundary detection or lane marking detection. In addition, to further improve the road or lane marking detection, tracking is employed to remove the erroneous road or lane information by considering the consecutive frames. The latter approach employs the neural network to learn the mapping function between the vision modality and the control actuator. For this purpose, vision data induced with the steering angle information is collected through behavior cloning (expert demonstrations) and utilized for training the neural network. Learning the mapping function through imitation learning involves neural network architecture selection, dataset collection, and optimizing the neural network parameters for better performance in predicting the steering angle. \cite{saleem2021steering}. 
\begin{figure}[t]
      \centering
      \includegraphics[width=11cm]{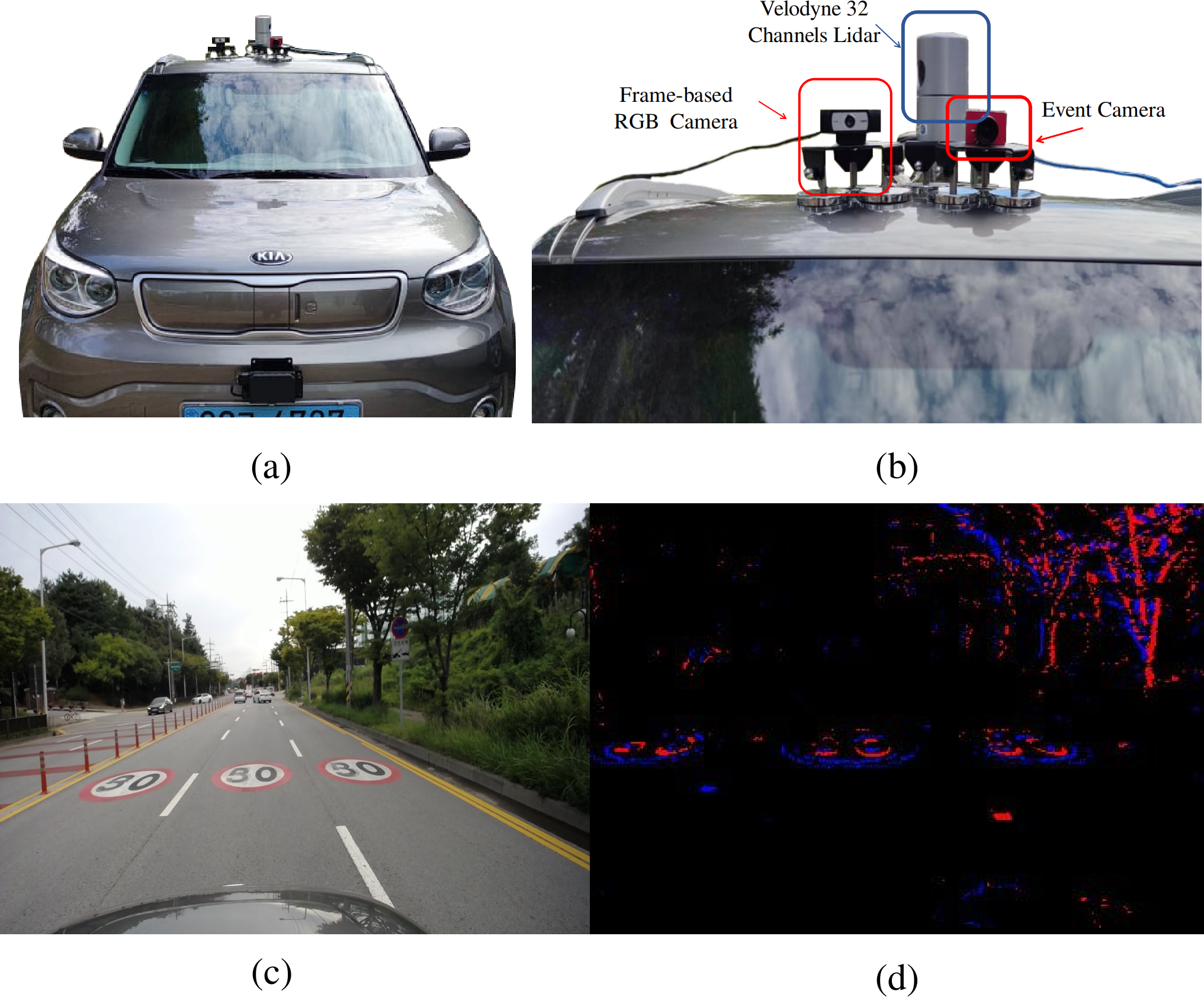}
      \caption{ (a,b) Our autonomous vehicle equipped with sensors for data collection. (c,d) shows the frame-based RGB image and event data captured using this setup.}
      \label{vehicle}
\end{figure}
\par 
However, the vision modality that is frame-based RGB cameras has shown impressive results for end-to-end driving \cite{codevilla2019exploring,chen2020learning,ohn2020learning,prakash2020exploring}. Yet, the robust performance of these vision systems for autonomous driving in challenging conditions is still an open problem. For instance, the frame-based camera has limitations in changing illumination conditions and prone to motion blur. However, the event camera captures the per-pixel brightness change as a stream of asynchronous events, providing complementary sensory information to the conventional frame-based RGB cameras.  The event camera operates in a high dynamic range and has high temporal resolution and less motion blur compared to frame-based RGB cameras \cite{gallego2019event}. 
\par 
Since the event camera provides complementary sensory information to the frame-based RGB camera in changing illumination conditions, this motivates the hypothesis of this work to explore the fusion mechanism of event and RGB frame-based cameras in the context of steering angle prediction for the lateral control. Prior work in the context of the event and frame-based cameras fusion involves two approaches. The first approach corresponds to augmenting the event as a channel to the frame-based RGB image \cite{hu2020ddd20}. In contrast, in the other approach, frames are generated from the event data for the feature representation learning \cite{maqueda2018event}. The earlier approaches have utilized the convolutional neural network as a feature extractor network with stacked event and frame-based RGB data that helps CNN capture the global context with a single modality. However, using this stacking nature limits the utility of the event camera as a secondary modality with the frame-based RGB images and leads to the non-trivial behavior of model interactions between multiple features.  
\par 
To overcome the problems mentioned earlier, this work explores learning the lateral control in terms of steering angle prediction by introducing the self-attention approach between the frame-based RGB and event data fusion. A convolutional encoder-decoder architecture named DRFuser is designed. In the proposed architecture (DRFuser), an individual convolutional encoder network is adopted for each of both modalities, including self-attention layers for the feature extraction and fusion, as illustrated in Fig.\ref{framework}. In this work, we follow the behavior cloning approach of imitation learning approach to learn the lateral control in terms of steering angle prediction. The efficacy of the proposed method is evaluated on three datasets, i) our collected dataset, ii) Davis Driving dataset (DDD) \cite{hu2020ddd20}, and iii) the simulated Carla EventScape dataset \cite{gehrig2021combining}. We have utilized our experimental vehicle as illustrated in Fig.\ref{vehicle} for the dataset collection installed with Davis-$346$ event camera and frame-based RGB camera \cite{azam2020system}. The vehicle control data is collected from the vehicle Controller Area Network (CAN) bus. In the experimental evaluation of the proposed method, Root Mean Square (RMSE) and Mean Absolute Error (MAE) metrics are employed, showing the better scores of the proposed method. As in the literature, the DDD dataset is only used for the experimental evaluation of predicting the steering angle; the proposed method has outperformed the state-of-the-art methods using the DDD dataset in terms of RMSE evaluation.
In summary, the main contributions of this work are as follows:
\begin{enumerate}
\item A novel convolutional encoder-decoder architecture (DRFuser) with the self-attention layer in each encoder's modality (event and frame-based RGB) is designed for learning end-to-end lateral steering angle. We explore the behavior cloning approach of imitation learning for learning the steering angle by fusing the event data and frame-based RGB data. 
\item We have collected our dataset comprises of event data, frame-based RGB data and Controller Area Network (CAN) bus data which include throttle, brake, speed , steering angle and torque values. All the data is collected using Robot Operating System (ROS) framework and synchronized using the ROS framework. Our code and data \footnote{\url{https://github.com/azamshoaib/DRFuser}} is open-sourced.
\item The efficacy of the proposed method is extensively evaluated on three datasets which include our collected dataset, DDD dataset and Carla EventScape dataset. We have evaluated the proposed method in term of RMSE and MAE scores for predicting the steering angle. 
\end{enumerate}

\section{Related Work}

\subsection{End-to-End lateral control learning for Autonomous vehicles}
End-to-end learning of lateral control in terms of steering angle prediction is a challenging problem.
The typical autonomous driving lateral control is developed by aggregating highly engineered modular architecture that includes localization, perception, motion planning, prediction, decision making, and control. In contrast, end-to-end approaches generate steering angle values directly mapped from visual observation to control actions. The visuomotor lateral control tightly couples the perception and control element of the problem.
 ALVINN was the first visuomotor prediction network that learned to steer from the images \cite{pomerleau1998autonomous}. The network consists of $3$ layers of a neural network trained in an end-to-end fashion. Therefore, ALVINN demonstrated the potential of neural networks for steering autonomous vehicles.  Following that, NVIDIA used the convolutional neural network (CNN) network to predict steering angles using images from the front-facing camera \cite{bojarski2016end}. They acquire good predictions on the highway driving by utilizing five convolution layers and three fully connected layers network. 
 
The researchers have used the studies mentioned earlier to introduce an assortment of networks to learn to steer. 
\cite{xu2017end} use an extensive video database to train the FCN-LSTM network to learn a generic vehicle motion model that predicts future steering angle from camera observation and previous vehicle state. \cite{kim2017interpretable} uses attention heat maps to regress steering angle from frame-based RGB. They emphasize the interpretability of input data to predict the control commands. \cite{azam2021n2c} leverage behavioral cloning to train a neural network-based controller which predicts throttle, brake, and torque commands using speed and steering values. Moreover, they have used images to predict speed and steering angle using a deep neural network which is used in conjunction with a neural network-based controller.

Moreover, some research is focused on multi-modal steering angle prediction. For example, \cite{maanpaa2021multimodal} uses RGB image and Lidar range and reflectance to predict steering angle for adverse weather conditions. The data is fused using two techniques; the middle fusion dual model and channel gated dual model. In the middle fusion, the features are concatenated after the convolution part of the network, whereas the channel gated dual model has two parallel architectures which concatenate the features through the gated channel. \cite{hou2019learning} supplement the visual input using auxiliary information from segmentation and optical flow data to learn the steering angle. They used PSPNet and FlowNet network to obtain segmentation and optical flow from RGB image and incorporate low, medium, and high level features map to train end-to-end network for steering angle prediction. 
\subsection{Dynamic vision sensor for visuomotor lateral control Autonomous vehicle}

Early work on end-to-end learning to steer used visual data from the frame-based camera. 
 However, we propose incorporating information fusion gathered from event-based asynchronous and frame-based cameras to learn the lateral control of the autonomous vehicle in an end-to-end learning manner. 
The event camera operates asynchronously and captures the change in brightness (events) for all pixels independently \cite{gallego2019event}. Therefore, the event camera can generate sparse asynchronous signals in space and time, enabling it to have a higher temporal resolution, low latency, and high dynamic range. The frame-based cameras are prone to illumination variation, motion blur, and sun glare. The complementary nature of event data to the RGB frame-based enhances it's applicability to be utilized for the fusion of information both the modalities. 
\par 
The proficiency of event cameras in providing rich data helps in solving perception problems in autonomous vehicles \cite{hidalgo2020learning,munir2021ldnet,alonso2019ev}. Here, we tackle the problem of steering angle prediction by fusing data from event-based cameras and frame-based RGB cameras. Similar work is done by \cite{maqueda2018event, hu2020ddd20}, they publish two large scale datasets, DDD$17$ and DDD$20$. These datasets contain event and RGB-frame data for different road and weather conditions. They designed a simple deep neural network based on ResNet-32 and linear layers to predict the steering angle. The input is a two-channel tensor consisting of event-frame and RGB-Frame. The network is trained on one sequence of recorded data and tested for the event-only data and fused data. In \cite{maqueda2018event} same network is used with different input data representation. The event data is converted to the frame using the integration of events with time. The proposed work is improved on \cite{maqueda2018event, hu2020ddd20} by developing a self-attention-based fusion network for steering angle prediction. Instead of using event data as an input channel, we explore the fusion of features by utilizing self-attention to emphasize valuable features and improve the network's learning.
\subsection{Attention}
The ability of attention to acquire to focus on essential features within a context has made it a significant ingredient in deep learning models for several modalities.
In literature, there are three kinds of attention used, additive \cite{bahdanau2014neural}, multiplicative \cite{luong2015effective} and self-attention \cite{vaswani2017attention}. The additive attention transfer the information from the encoder to the decoder, which enhances the feature representation. The decoder neuron receives additional input through a gating signal from the encoder providing flexibility to focus on essential features. However, in multiplication attention, the gating signal is multiplied instead of addition, and it has the drawback of poor performance for high dimensional input features. 
The self-attention incorporates long-distance interactions into the model, which gives it the strength to remember global and local features. We incorporate a self-attention module to fuse the information from the event and frame-based data to encode the surrounding environment's global context, which helps the algorithm predict the steering angle robustly. 
\begin{figure}[t]
      \centering
      \includegraphics[width=11cm]{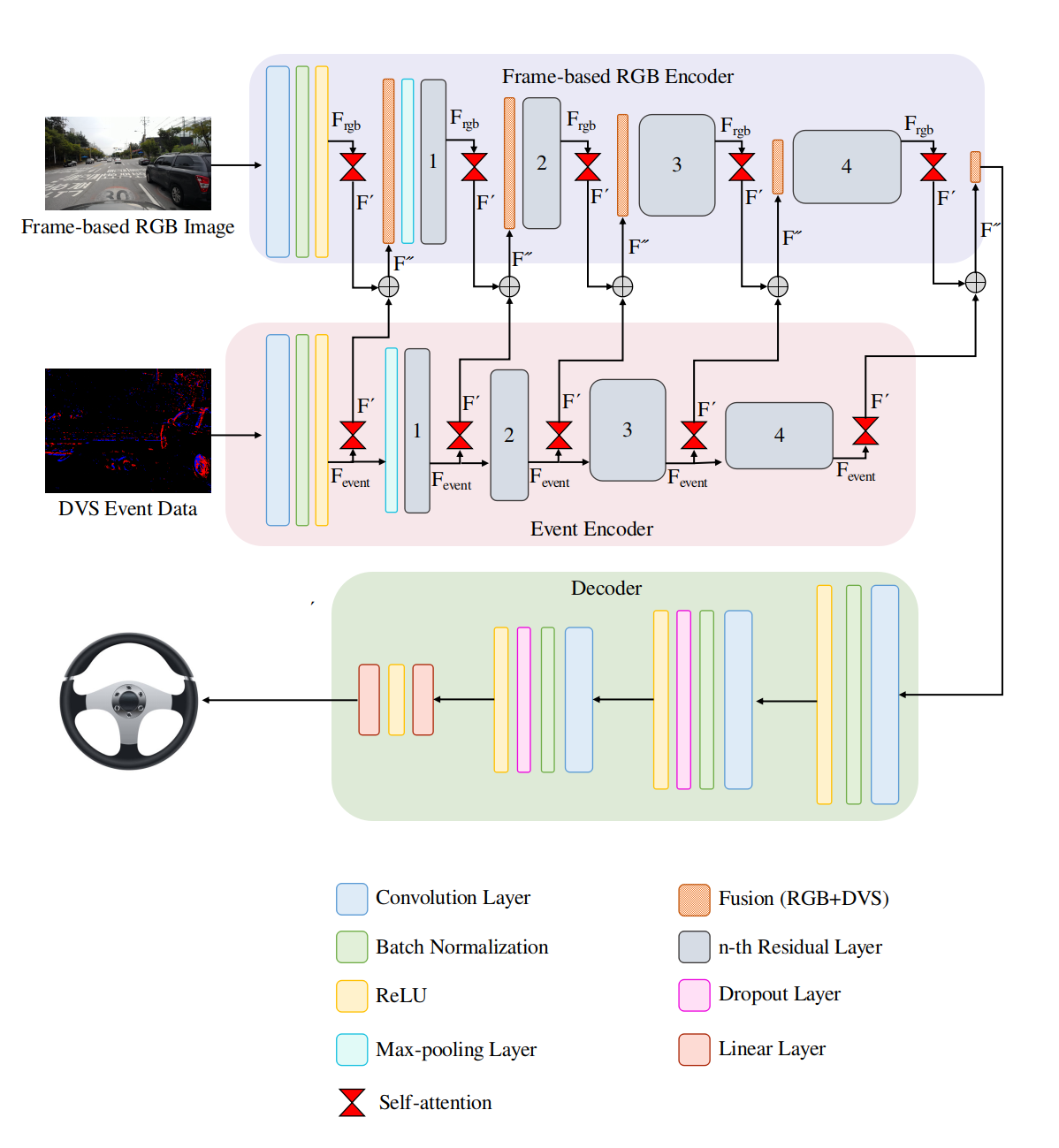}
      \caption{The overall framework for the proposed method. It includes convolution encoders, self-attention layer for feature extraction and fusion and a convolution decoder. The network input frame-based RGB and event data and predict steering angle. }
      \label{framework}
\end{figure}

\section{Methodology}
This section elaborates in detail a novel proposed framework for learning end-to-end lateral control steering angle from multi-modal data, as illustrated in Fig.2. The framework is composed of the convolutional encoders for frame-based RGB images and event data, self-attention layers and a convolutional decoder. The self-attention layers are introduced between the encoders to learn the long-range interaction between the encoded features for frame-based RGB image and event data. 
\subsection{Problem Formulation}
This work explores end-to-end learning of the lateral control by predicting steering angle in an urban setting through fusing the information from frame-based RGB and event data.
% This work explores end-to-end learning the control policy in the form of steering angle prediction in an urban setting by fusing the information from frame-based RGB and event data. 
\par 
Imitation learning (IL) deals with learning a control strategy $\pi$ that demonstrates the behaviour of an expert ${\pi}^*$ \cite{hussein2017imitation}. Employing the IL strategy, the focus of our problem is to learn the lateral control by mapping the input to the steering angle. We have adopted IL's Behavior Cloning (BC) approach that corresponds to the supervised learning method \cite{ly2020learning}. An expert behavior cloning dataset $D = {(X^i,S^i)_{i}^{T}}$ of size $T$ is collected from the environment. The dataset $D$ comprises of high-dimensional observations $X$ with corresponding steering angle $S$ recorded by manual driving. The fusion network in the form of encoder-decoder architecture is trained in a supervised manner using the dataset $D$ with the loss function $\mathfrak{L}$ by employing the objective function as expressed as: %in Eq.\ref{equ1}
\begin{flalign}
\label{equ1}
\hspace{-1cm}
    \begin{aligned}
   \arg \min _{\pi}\mathbb{E}_{(X,S)\rightarrow D}[\mathfrak{L}(P,\pi(X))].
    \end{aligned}
\end{flalign}
The high-dimensional observation, $X$, encompasses the frame-based RGB image and event camera data. In training, the Huber loss (smooth) $L_1$ loss function is employed to measure the distance between the predicted steering angle, $\pi(X)$ and the expert steering angle $P$. We have used Huber loss (smooth-L1 loss) as a loss function given below, the  $\delta=1.0$ is used in the experimentation. The Huber loss function is combination mean squared error and the absolute value function. The intuition to use Huber loss as the objective function in training the proposed method is to combine the best of two world (i-e mean squared error and absolute value function). The balancing nature of Huber loss between mean squared error and mean absolute error allow to train the proposed model for varied data (as in the case of steering angle). The value of $\delta$ in the Huber loss determines the applicability of mean squared error and mean absolute error as an objective function in training the proposed network. For the loss values that are less than $\delta$, mean squared error is employed whereas for larger loss values in comparison to $\delta$ mean absolute error is employed.
\begin{flalign}
\label{eq8}
\hspace{2.7 cm}
L_{\delta}(x) = 
\left\{\begin{matrix}
\frac{1}{2}x^{2} \;&&for \abs{ x}  \leq \delta \\
\delta(\abs{ x}  - \frac{1}{2}\delta ) \;&& otherwise 
\end{matrix}\right. \;  &&
\end{flalign}
% L_{\delta}(x) = 
% \left\{\begin{matrix}
% \frac{1}{2}x^{2} \;&&for \abs{ x}  \leq \delta \\
% \delta(\abs{ x}  - \frac{1}{2}\delta ), \;&& otherwise 
% \end{matrix}\right. \;  &&

\subsection{Self-attention Multi-Modal Fusion Network}
The convolution in deep neural networks forms a fundamental building block for current vision architectures. However, convolution is unable to acquire long-term dependencies of the features, and effort is focused on augmenting convolution modules with non-local means such as attention to gain advancement on vision tasks. 
The introduction of attention in the encoder-decoder architecture for the neural transduction models has enabled learning the representation from variable sources  \cite{bahdanau2014neural}. Besides, the success of attention in natural language processing applications approaches incorporating attention to vision tasks have illustrated tremendous performance. Specifically, self-attention is in the form of attention applied within a single context instead of across multiple contexts. Here, context can be regarded as different modalities, for instance, speech, text, or images \cite{xu2015show,wu2016google}. The capability of attention to learn to concentrate on essential features within the context has motivated this study to use it in the fusion of multi-modal data. 
% The introduction of attention in the encoder-decoder architecture for the neural transduction models has enabled learning the representation from variable source sentences \cite{bahdanau2014neural}. Besides, the success of attention in natural language processing applications approaches incorporating attention to vision tasks have illustrated tremendous performance. Specifically, self-attention is in the form of attention applied within a single context instead of across multiple contexts. Here, context can be regarded as different modalities, for instance, speech, text or images \cite{xu2015show,wu2016google}.
\par
The self-attention multi-modal fusion network comprises encoder-decoder architecture with the addition of self-attention layers in the encoder module. The key idea is to exploit the self-attention layer to incorporate the global context of frame-based RGB image and event data in predicting steering angle. Formally, given the input features $\mathbf{F}_{ij} \in \mathbf{R}^{C_{in}}$, where $C^{in}$ is the number of channels that correspond to the input features. A local region of features known as memory bank in the position of $uv \in N_k(i,j)$ with the spatial extent of $k$ centred around $\mathbf{F}_{ij}$ is extracted for the computation of self-attention. This local form of self-attention is disparate from the global attention between all features. Global attention is computationally expensive, which limits its applicability across all layers in the encoder-decoder network. The linear projections for computing the set of queries, keys and values ($Q_{ij}$,$K_{uv}$ and $V_{uv}$) are expressed in Eq.(\ref{equ2}) \cite{vaswani2017attention}.
\begin{flalign}
\label{equ2}
\hspace{-1cm}
    \begin{aligned}
   \mathbf{Q}_{ij}= \mathbf{W}_{Q}\mathbf{F}_{ij}, \mathbf{K}_{uv}=\mathbf{W}_{K}\mathbf{F}_{uv}, \mathbf{V}_{uv}=\mathbf{W}_{V}\mathbf{F}_{uv},
    \end{aligned}
\end{flalign}
where $ \mathbf{W}_{Q}, \mathbf{W}_{K}, \mathbf{W}_{V} \in \mathbb{R}^{C_{out} \times C_{in}}$ are the learned weight matrices. The single-headed attention is computed by employing the dot-product between $\mathbf{Q}_{ij}$ and $\mathbf{K}_{uv}$ and then aggregates the values $\mathbf{V}_{uv}$ for each query as expressed in Eq.(\ref{equ3})
\begin{flalign}
\label{equ3}
\hspace{-1cm}
    \begin{aligned}
{F}' = \sum_{u,v \in N_k(i,j)} \underset{uv}{softmax}( \mathbf{Q}_{ij}^T\mathbf{K}_{uv})\mathbf{V}_{uv},
    \end{aligned}
\end{flalign}
where $\mathbf{F'}$ corresponds to the output feature of same size as the input feature $\mathbf{F}$. In practice, multiple attention head are employed, so the aforementioned computation is performed repeatedly for every feature $(ij)$. For the computation efficacy, the input feature $\mathbf{F}_{ij}$ is partitioned in to $M$ groups and single-head attention is computed for each group with different learned weight matrices per head, and finally concatenated to give the output representations. To incorporate the positional information in the attention for permutation inequivariant, relative distance of $ij$ is added to each position of $uv \in N_k(i,j)$. This relative distance is offseted by row and column offset as denoted as $u-i$ and $v-j$ respectively. In addition, these offsets are associated with embedding $e_{u-i}$ and $e_{v-j}$ respectively. The relative attention that is employed in the feature extractor is expressed in Eq.(\ref{equ4}) \cite{vaswani2017attention}.
\begin{flalign}
\label{equ4}
\hspace{-1cm}
    \begin{aligned}
{F}'=\sum_{u,v \in N_k(i,j)} \underset{uv}{softmax}(\mathbf{Q}_{ij}^T\mathbf{K}_{uv}+\mathbf{Q}_{ij}^T\mathbf{e}_{u-i,v-j})\mathbf{V}_{uv}.
    \end{aligned}
\end{flalign}
\par 
For feature extraction and fusion, two encoder networks are designed for frame-based RGB image and event data, respectively. The architecture of both encoder networks are identical to each other. In contrast to SegNet \cite{badrinarayanan2017segnet}, MFNet \cite{ha2017mfnet} and FuseNet \cite{hazirbas2016fusenet}, we have employed the ResNet \cite{he2016deep} model for the feature extraction. The last two layers of ResNet consisting of 
 average pooling and fully connected layers are excluded from the ResNet model as they were used for the classification task. ResNet model incorporates an initial block containing convolution, batch normalization, and ReLU activation layer. A max-pooling layer and four residual blocks are sequentially used after the initial block to reduce the spatial resolution of the features. Here it is to be mentioned that in this work, we have the ResNet model as it is highly accepted as a feature extractor in the research community. 
\par 
The self-attention layer is employed in each encoder to learn the long-range interaction between the encoded features. The self-attention layers are applied at multiple scales in the ResNet feature extraction, as illustrated in Fig.\ref{framework}. The output of the self-attention layer from both encoders networks is fused element-wisely. It is to be noted that the feature map size is not changed after the fusion. The resulting output from the last fusion operation is fed as input to the decoder network.
\par 
The function of the decoder is to predict the prediction based on the encoded features. We have designed a simple yet effective decoder for steering angle prediction in the proposed work. The decoder is not the mirrored version of the encoder network. The encoder and decoder are asymmetric in architecture. The decoder network comprises three convolutional blocks; each includes convolution, batch normalization, and ReLU activation layer. In the second and third convolutional block, a dropout layer is introduced for generalization and avoiding the over-fitting of the model. Followed by convolution blocks, two linear layers are employed to predict steering angle. The decoder architecture details are given in Table-\ref{decoder}. The decoder used with ResNet-34 has one convolution block compared to ResNet-50 to accommodate the feature map size. 

\begin{figure}[t]
      \centering
      \includegraphics[width=9cm]{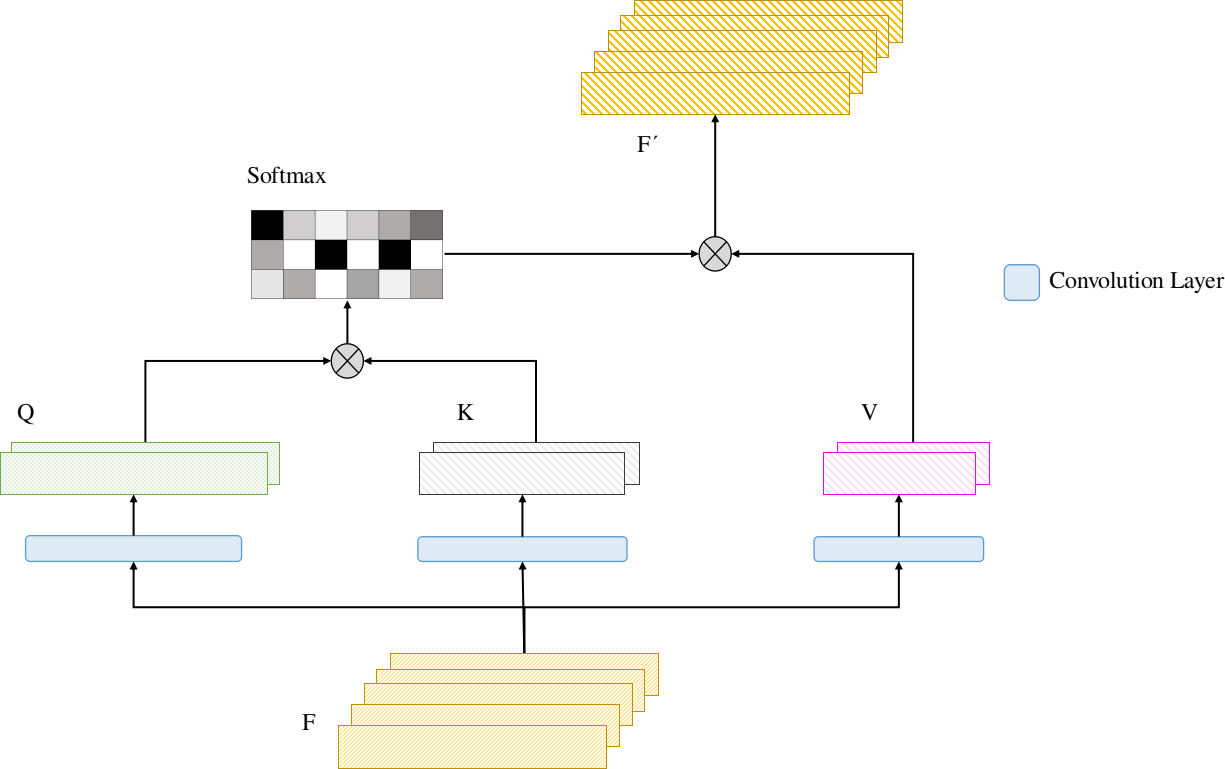}
      \caption{The detail of working operation of self-attention layer is illustrated.}
      \label{flir}
\end{figure}

\section{Experimentation and Results}

\subsection{Datasets}
The efficacy of the proposed method for learning the control policies using the fusion of frame-based RGB image and event data is evaluated on two available public datasets and our collected dataset. The details of each dataset is explained below. 
% The effectiveness of predicting control policies by fusing event data and frame-based RGB images is evaluated on two public datasets and our collected dataset. 
\subsubsection{Our Collected dataset}
We have used our experimental vehicle shown in Fig.\ref{vehicle} equipped with exteroceptive and proprioceptive sensors for data collection. The perception sensor includes $32$-channels Velodyne Lidar, Logitech RGB camera and DAVIS$346$ dynamic vision event camera. In addition, the Novatel Global Navigation Satellite System (GNSS) is incorporated as the proprioceptive sensor. Finally, the drivekit\footnote{\url{https://github.com/PolySync/oscc}} and can-shield are used to extract the CAN bus data. The details of which are given in our previous work \cite{azam2021n2c}.
This study used CAN bus data, frame-based RGB image data, and event data for the proposed method. The event data is collected using a DAVIS$346$ camera, which has a resolution of $346 \times 260$ pixels. The DAVIS$346$ camera also outputs an intensity image, but the resolution quality of that intensity image is low. To compensate for this, we have utilized the Logitech-c$920$ RGB camera. It gives a max resolution of $1080$ pixel at $30$ frame per second (fps). Meanwhile, the CAN data compromises torque, speed, throttle, brake, and steering angle are obtained through behavioural cloning while driving in an urban environment during day and night. 
\par 
The event camera, frame-based RGB camera and vehicle CAN are all operating at different frequencies. The difference in operating frequencies limits the utilization of this data for the proposed method. For this purpose, the data from the sensors, as mentioned earlier, need synchronizing. Generally, there are two synchronizing approaches adopted in the research community that are hardware-triggered and soft-time synchronization.  In the proposed method of data collection, the soft-time synchronization approach is adopted. The soft-time synchronization approach is developed under the Robot Operating System (ROS)\footnote{\url{https://www.ros.org/}} framework. ROS provides a flexible ecosystem for robotics systems and contains powerful tools and libraries. Therefore, for the data synchronization, we have utilized the ROS framework.  We have employed individual ROS nodes for the data acquisition from CAN bus, frame-based RGB camera and event camera. We have designed a proprietary ROS node for the CAN bus that publishes speed, steering angle, brake, torque, and throttle values in the ROS framework. Since ROS provides drivers for the frame-based RGB camera, we have utilized that ROS node for the RGB image data collection. For the DAVIS camera, we have employed the open-source ROS node for event data collection. Since each sensor has a different frequency, we adopted the nearest neighbour search based on the lowest sensors' frequency. We matched the corresponding other sensor data with that for the data synchronization. In our case, the frame-based RGB camera has the lowest frequency, whereas the event camera and CAN bus have high frequencies. In addition, we have open-sourced the synchronization code for the research community.

\paragraph*{Pre-processing of Event Data}
% \subsubsection{Pre-processing of Event Data}
Event cameras are operated asynchronously and generate output data in the form of spikes or events; as a result, from the change in brightness level in the viewing scene. The event or spike representation of event camera data limits its usability for the convolutional neural network. For this purpose, the event data is transformed to a suitable representation (for instance, images) that the convolutional neural networks can employ. Mathematically, to convert a stream of event data to an image where each independent pixel corresponds to a change in brightness $\mathbf{E}(\mathbf{v_i},t_i)\doteq log \mathbf{I}(\mathbf{v_i},t_i)$. An event is recorded for each pixel $\mathbf{v_i} = (x_i,y_i)^T$ and time $t_i$ when there is a change in brightness and surpasses the threshold ($\eta$) is expressed as :% Eq.\ref{equ5} \cite{gallego2019event}.
\begin{flalign}
\label{equ5}
\hspace{-1cm}
    \begin{aligned}
\mathbf{E}(\mathbf{v_i},t_i)-\mathbf{E}(\mathbf{v_i},t_i-\Delta t_i)> p_i\eta ,
    \end{aligned}
\end{flalign}
where $p_i \in {-1,1}$ represent polarity of brightness and $\Delta t_i$ corresponds to the time of last triggered event at location $v_i$. Eq.(\ref{equ5}) represents the event generation model and using this a sequence of events $\mathbf{E}(t_N)= e_{i_i=1}^{N}=(x_i,y_i,t_i,p_i)$ is generated for the time interval $\Delta t_i$. In our data collection, we fixed the events to $100000$, which is approximately $50$msec and generate the corresponding image representation when the number of events reached to specific number as mentioned earlier. The number of events selection is empirical and produce better results in our data collection. 

\par
We have collected $297.3$ GB of raw driving data in an urban environment at day and night, respectively. We have processed $110$ GB of data for our experimentation by converting event data into image representation and synchronizing it with frame-based RGB and vehicle control data. The processed data consists of $71274$ image pairs of event and frame-based RGB and steering angle information, split into training ($33764$) and testing ($37510$) sets. The data is split carefully, consecutive and non-overlapping sequences of data is considered for training and test sets.
\subsubsection{Davis Driving Dataset (DDD)}
The Davis Driving Dataset (DDD) is one of the most extensive datasets collected using a DAVIS camera, a dynamic and active pixel vision sensor \cite{hu2020ddd20}. It provides a concurrent stream of events and an active pixel sensor (standard grayscale images). The events represent brightness changes occurring at a particular moment and have a dynamic range of $120$dB and an effective frame rate of $1kHz$. The DAVIS camera has a resolution of $346 \times 260$ pixels. The dataset comprises DAVIS camera data (frames and events), and vehicle human control data in the form of speed and steering angle is collected for approximately $51$ hours of driving on urban and highway roads under different weather and illumination conditions. The total data size is approximately $1.3TB$, from which we have sampled  $63.7GB$ of data containing different weather and lighting conditions as shown in Table-\ref{table-DDD}.
The consecutive and non-overlapping recordings are split into training and test set. The data is pre-processed on the same principle as explained above. The events are accumulated for the time interval of $50$msec and represented as an image. The training set contains $236566$ image pairs and test set $54303$.
% \usepackage{booktabs}
% \usepackage{graphicx}
\begin{table}[t]
\centering
\caption{The list of sample data included in training and testing from Davis driving data.}
\label{table-DDD}
\resizebox{8cm}{!}{%
\begin{tabular}{@{}lcccc@{}}
\toprule
Filename   & Scene   & Condition & T(s) & GB   \\ \midrule
1487354030 & City    & night,wet & 377  & 3    \\
1487354811 & City    & night,wet & 190  & 1.4  \\
1487417411 & Freeway & day       & 2096 & 18.2 \\
1487597945 & City    & evening   & 50   & 0.5  \\
1487598202 & freeway & day       & 1882 & 15.1 \\
1487778564 & campus  & day       & 101  & 1.1  \\
1487839456 & City    & day,sun   & 406  & 5.7  \\
1487849151 & town    & day,sun   & 429  & 5.5  \\
1487856408 & town    & day,sun   & 817  & 13.2 \\ \bottomrule
\end{tabular}%
}
\end{table}
\subsubsection{EventScape}
EventScape is a large-scale synthetic dataset recorded through the CARLA simulator \cite{gehrig2021combining}. An event camera sensor is implemented in CARLA simulator \cite{dosovitskiy2017carla}, which renders images by computing per-pixel brightness change from the simulated environment. The similar pre-processing method is adopted as explained above. The event data for an approximate time interval of 2ms is rendered into a single image frame. The camera parameters are set to $512 \times 256$ resolution and focal length of $256$ pixels.  In addition, the simulated sensor can record various environmental conditions, such as different weather conditions and illumination variations. 
In this work, we have utilized the data published by \cite{gehrig2021combining}. It consists of approximately $2$ hours of driving in different towns and weather conditions. The data contains event data frames, frame-based RGB images, and vehicle control data. The training set consists of data collected from Town-$1$, $2$, and $3$, while validation and test data are collected from Town-$5$. The training set contains $122330$ image pairs, and the test set has $26182$.

\begin{figure}[t]
      \centering
      \includegraphics[width=9cm]{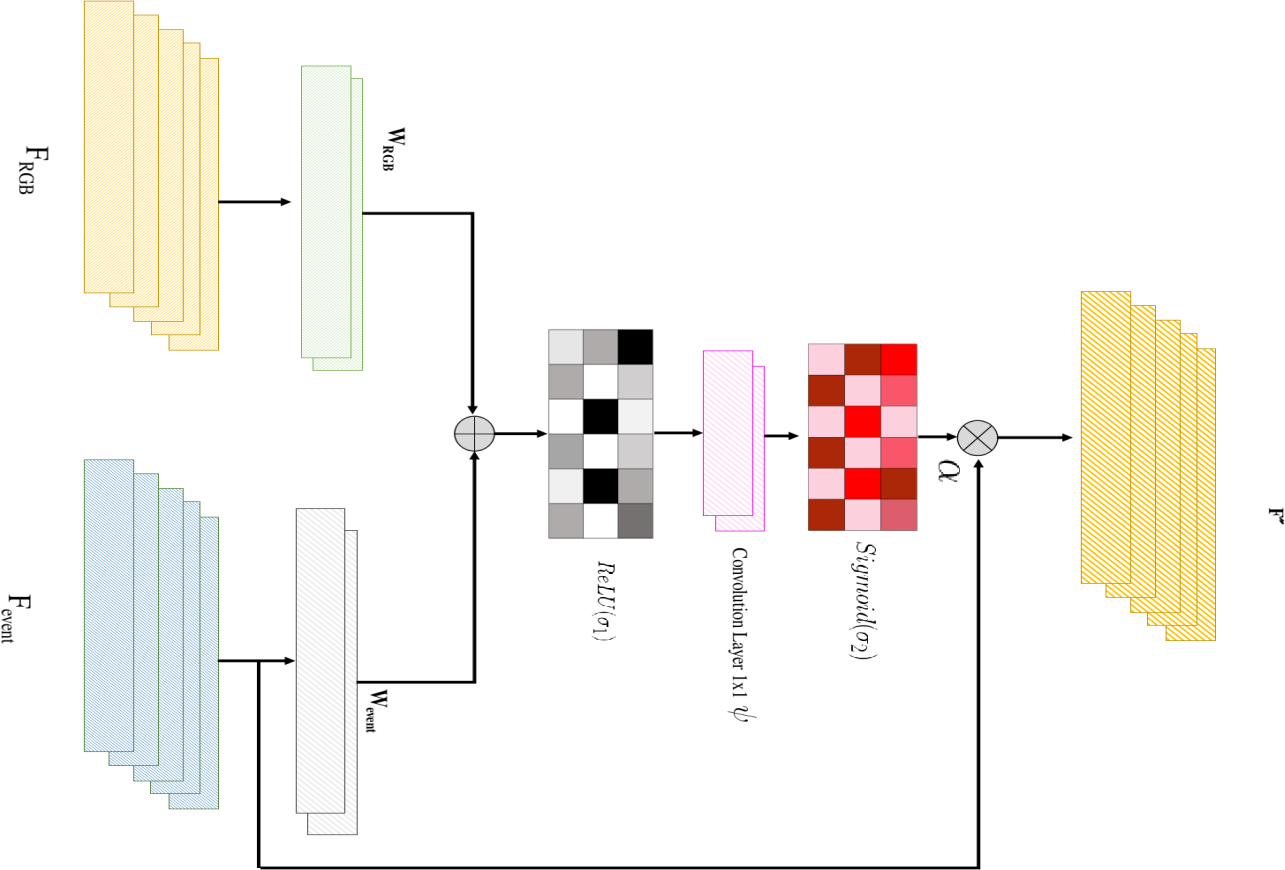}
      \caption{The additive attention layer mechanism is shown, where it input two feature set from each encoder and compute fusion of features. }
      \label{additive-attention}
\end{figure}
\subsection{Baseline}
Towards gaining a deep insight into our proposed network, which fuse the event and frame-based RGB by employing a self-attention layer in the encoder. Here we investigate the effect of the self-attention layer on fusing the event and frame-based RGB by performing experiments using the No-attention layer and the additive attention layer \cite{bahdanau2014neural}. The No-attention layer is simply element-wise addition of features maps obtained from the event encoder and RGB encoder, as shown by  Eq.(\ref{equ45}). 
\begin{flalign}
\label{equ45}
\hspace{-1cm}
    \begin{aligned}
{F}''={F_{RGB}}+{F_{event}}.
    \end{aligned}
\end{flalign}
For additive attention layer utilizes an additive attention mechanism to fuse data between event and RGB encoder.  The Fig.\ref{additive-attention} illustrates the working principle of additive attention. The attention coefficients $\alpha_{i} $ identify the salient regions in the frame-based RGB and prune the relevant features, which are fused with event data by element-wise multiplication. The output of the additive attention layer is ${F_i}''={F}_{event,i}\cdot \alpha_i$. The  $\alpha_{i} $ is given by Eq.(\ref{eq5}) \cite{bahdanau2014neural}.
\begin{flalign}
% \hspace{-1cm}
% \hspace{-2.5cm}
\begin{aligned}
& h_{att}=\psi^T (\sigma (W^T_{RGB}{F}_{RGB,i}+W^T_{event}{F}_{event,i}))+b_\psi, \\
\end{aligned}
\end{flalign}
\begin{flalign}
% \hspace{-1cm}
% \hspace{-2.5cm}
\begin{aligned}
 & \alpha _i=\sigma _2(h_{att}({F}_{RGB,i},{F}_{event,i};\Theta_{att} )). 
 \label{eq5}
\end{aligned}
\end{flalign}
where $\sigma _2({F}_{i})=\frac{1}{1+exp(-{F}_{i,})}$ characterised sigmoid activation function. $\Theta_{att} $  represents set of learning parameters $W_{RGB}$ , $W_{event}$, $\psi$ and bias term, $b_g$.These parameters are trained with standard back-propagation.

\begin{figure}[]
      \centering
      \includegraphics[width=10cm]{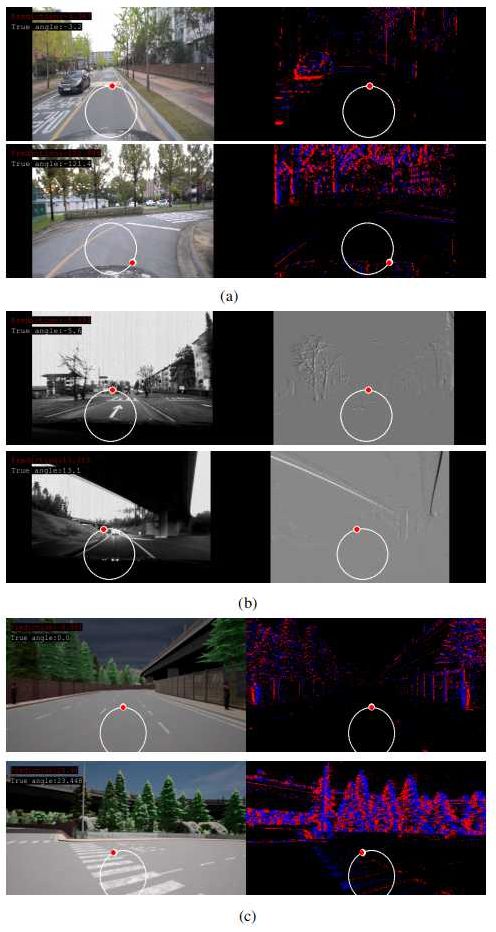}
      \caption{The visualization of prediction of steering angle in comparison of ground-truth, (a) shows collected dataset, (b) shows Davis driving data  and (c) shows Carla EventScape dataset. }
      \label{visualization}
\end{figure}
\begin{figure}[]
      \centering
      \includegraphics[height=11cm,angle=-90]{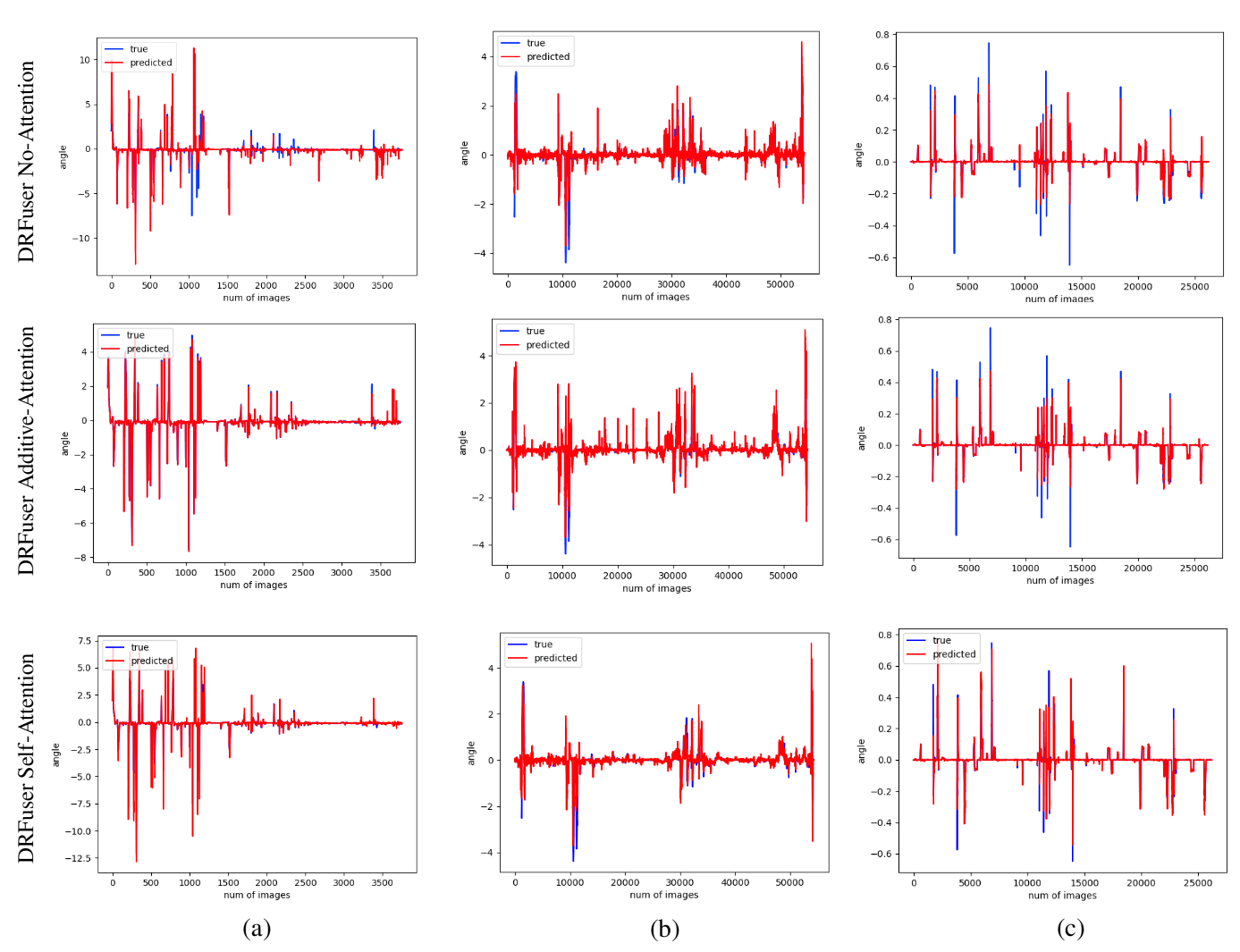}
      \caption{The quantitative comparison between the proposed method self-attention DRFuser, DRFuser with No-attention and additive attention layer on test data is illustrated. (a) Steering angle prediction on our collected dataset. (b) Steering angle prediction on Davis driving data, and (c) Steering angle prediction on Carla Eventscape dataset is presented. }
      \label{quantitative-proposed}
\end{figure}

\begin{figure}[]
      \centering
      \includegraphics[height=10cm,angle=-90]{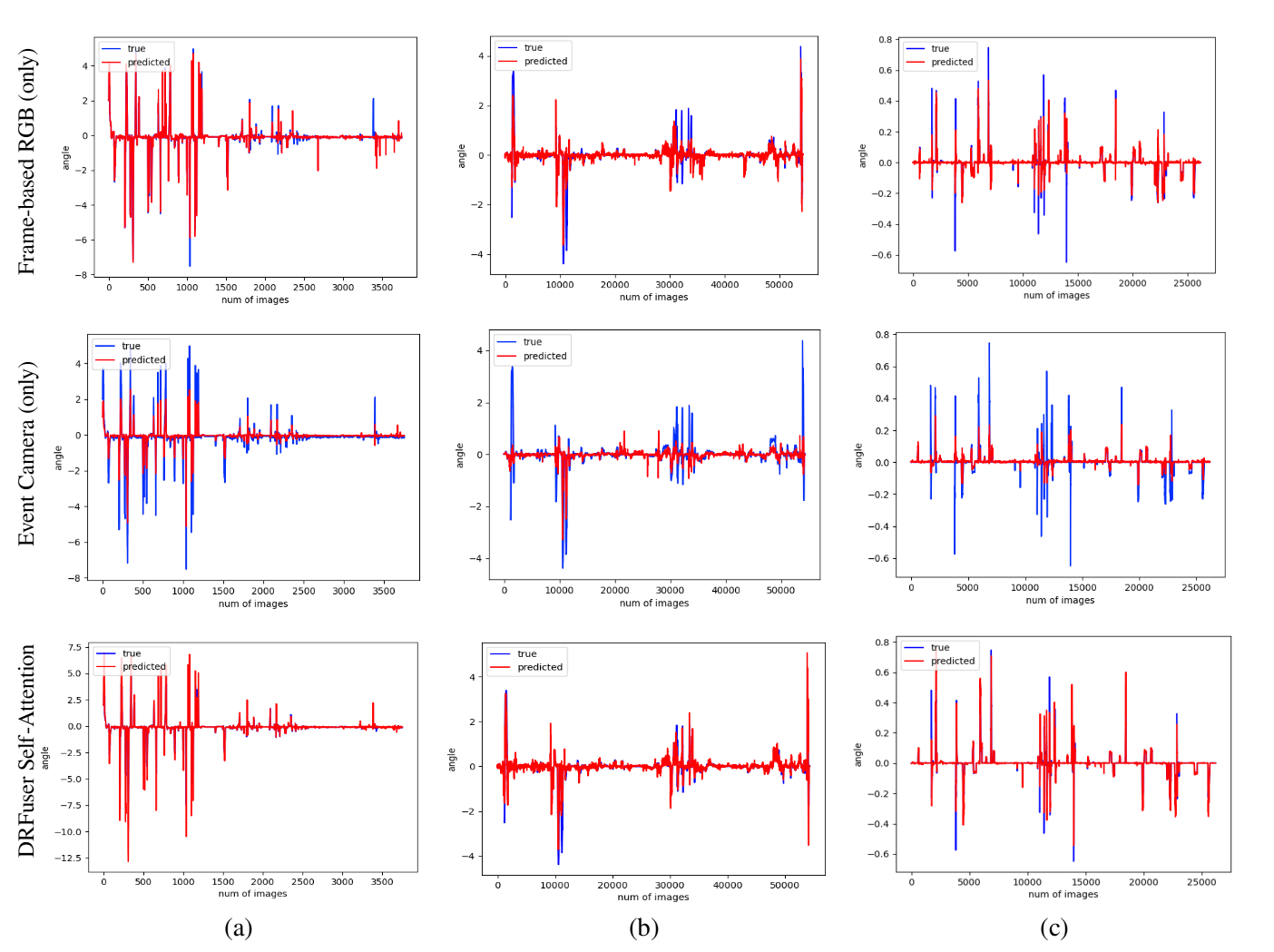}
      \caption{The quantitative comparison between frame-based RGB and event data fusion using the proposed framework DRFuser self-Attention (ResNet-50) and ResNet-50 trained on singular modality is shown. (a) shows steering angle prediction on our collected dataset. (b) illustrate the steering angle prediction on Davis driving data and (c) shows the steering angle prediction on Carla Eventscape dataset.}
      \label{quantitative-comp}
\end{figure}

\subsection{Training Details}
We have implemented the proposed network DRFuser using PyTorch 1.8.0 with CUDA 11.1. The DRFuser is trained on a PC with Intel i$7$ core and RTX $3090$ graphic card. Since the graphic card memory determines the batch size of the training network, it was changed accordingly for different networks. The minimum batch size of 2 was employed.
The network is trained end-to-end using pre-trained ImageNet weights of ResNet.  The image size of frame-based RGB and event data are identical for input to the network. We have used AdamW optimization solver for training. The weight decay, learning rate and momentum are set to $0.002$, $0.0001$ and $0.9$, respectively. The network is trained for 100 epochs.  
\par
% We have used Huber loss (smooth-L1 loss) as a loss function given below, the  $\delta=1.0$ is used in the experimentation. 
% \begin{flalign}
% \label{eq8}
% \hspace{0.7cm}
% L_{\delta}(x) = 
% \left\{\begin{matrix}
% \frac{1}{2}x^{2} \;&&for \left | x \right | \leq \delta \\
% \delta(\left | x \right | - \frac{1}{2}\delta ), \;&& otherwise 
% \end{matrix}\right. \;  &&
% \end{flalign}
Two evaluation metrics, root means square error (RMSE) and means absolute error (MAE), are used to determine the effectiveness of the proposed network, as shown by Eq.(\ref{eq9}) and Eq.(\ref{eq99})
\begin{flalign}
\label{eq99}
\begin{aligned}
\hspace{-0.7cm}
 & RMSE = \sqrt{\frac{1}{k}\sum_{j=1}^{k}(y_k-\hat{y_k})^2},\; \\
 \end{aligned}
\end{flalign}
 \begin{flalign}
\label{eq9}
\begin{aligned}
\hspace{-0.7cm}
& MAE = \frac{1}{k}\sum_{j=1}^{k}\left \| y_k-\hat{y_k} \right \|. \; 
\end{aligned}
\end{flalign}

% \usepackage{booktabs}
% \usepackage{graphicx}
\begin{table}[t]
\centering
\caption{Root Mean Squared Error (RMSE) and Mean Absolute Error (MAE) Scores for Our Collected Dataset for the proposed method and baseline methods.}
\label{Table-1}
\resizebox{12cm}{!}{%
\begin{tabular}{@{}lcccc@{}}
\toprule
Model                                   & \multicolumn{1}{l}{Frame-based RGB} & \multicolumn{1}{l}{Event Data} & RMSE Score & MAE Score \\ \midrule
ResNet-34                               & \cmark               & \xmark          & 0.2396     & 0.1295    \\
ResNet-34                               & \xmark               & \cmark          & 0.3234     & 0.1913    \\
ResNet-50                               & \cmark               & \xmark          & 0.2046     & 0.1143    \\
ResNet-50                               & \xmark               & \cmark          & 0.2782     & 0.1698    \\
No attention (ResNet-34 backbone)       & \cmark               & \cmark          & 0.1837     & 0.0963    \\
No attention (ResNet-50 backbone)       & \cmark               & \cmark          & 0.1773     & 0.0842    \\
Additive attention (ResNet-34 backbone) & \cmark               & \cmark          & 0.1722     & 0.0808    \\
Additive attention (ResNet-50 backbone) & \cmark               & \cmark          & 0.1612     & 0.0713    \\
DrFuser (ResNet-34 backbone)            & \cmark               & \cmark          & 0.1430     & 0.0540    \\
DrFuser (ResNet-50 backbone) & \cmark & \cmark & \textbf{0.1266} & \textbf{0.0396} \\ \bottomrule
\end{tabular}%
}
\end{table}

% \usepackage{booktabs}
% \usepackage{graphicx}
\begin{table}[!b]
\centering
\caption{Root Mean Squared Error (RMSE) and Mean Absolute Error (MAE) Scores for EventScape Dataset for the proposed method and baseline methods.}
\label{Table-2}
\resizebox{12cm}{!}{%
\begin{tabular}{@{}lcccc@{}}
\toprule
Model                                   & \multicolumn{1}{l}{Frame-based RGB} & \multicolumn{1}{l}{Event Data} & RMSE Score      & MAE Score        \\ \midrule
ResNet-34                         & \cmark & \xmark & 0.0293 & 0.0093  \\
ResNet-34                         & \xmark & \cmark & 0.0562 & 0.0134  \\
ResNet-50                         & \cmark & \xmark & 0.0281 & 0.0082  \\
ResNet-50                         & \xmark & \cmark & 0.0475 & 0.0122  \\
No attention (ResNet-34 backbone) & \cmark & \cmark & 0.0191 & 0.0051  \\
No attention (ResNet-50 backbone) & \cmark & \cmark & 0.0174 & 0.0049  \\
Additive attention (ResNet-34 backbone) & \cmark               & \cmark          & 0.0169          & 0.00435          \\
Additive attention (ResNet-50 backbone) & \cmark               & \cmark          & 0.0148          & 0.0040           \\
DrFuser (ResNet-34 backbone)      & \cmark & \cmark & 0.0125 & 0.00297 \\
DrFuser (ResNet-50 backbone)            & \cmark               & \cmark          & \textbf{0.0118} & \textbf{0.00214} \\ \bottomrule
\end{tabular}%
}
\end{table}

% \usepackage{booktabs}
% \usepackage{graphicx}
\begin{table}[t]
\centering
\caption{Root Mean Squared Error (RMSE) and Mean Absolute Error (MAE) Scores for Davis Driving Dataset (DDD) for the proposed method and baseline methods. }
\label{Table-3}
\resizebox{12cm}{!}{%
\begin{tabular}{@{}lcccc@{}}
\toprule
Model                                   & \multicolumn{1}{l}{Frame-based RGB} & \multicolumn{1}{l}{Event Data} & RMSE Score & MAE Score \\ \midrule
ResNet-34                               & \cmark               & \xmark          & 0.03363     & 0.01107    \\
ResNet-34                               & \xmark               & \cmark          & 0.03749     & 0.01312    \\
ResNet-50                               & \cmark               & \xmark          & 0.03242     & 0.01009    \\
ResNet-50                               & \xmark               & \cmark          & 0.03582     & 0.01211    \\
No attention (ResNet-34 backbone)       & \cmark               & \cmark          & 0.03023     & 0.00988    \\
No attention (ResNet-50 backbone)       & \cmark               & \cmark          & 0.02992     & 0.00934    \\
Additive attention (ResNet-34 backbone) & \cmark               & \cmark          & 0.02355     & 0.00812    \\
Additive attention (ResNet-50 backbone) & \cmark               & \cmark          & 0.02118     & 0.00798    \\
DrFuser (ResNet-34 backbone)            & \cmark               & \cmark          & 0.01654     & 0.00721    \\
DrFuser (ResNet-50 backbone) & \cmark & \cmark & \textbf{0.01519} & \textbf{0.00631} \\ \bottomrule
\end{tabular}%
}
\end{table}

% \usepackage{booktabs}
% \usepackage{graphicx}
\begin{table}[]
\centering
\caption{The comparison  of the evaluation  result of DRFuser  with other state-of-the-art method on Davis Driving Dataset. All RMSE scores are in radians. }
\label{Table-4}
\resizebox{12cm}{!}{%
\begin{tabular}{@{}lccc@{}}
\toprule
Model           &Frame-based RGB                   & Event Data                & RMSE Score      \\ \midrule
Bojarski et al. \cite{bojarski2016end} & \cmark & \xmark & 0.1574287 \\
CNN-LSTM \cite{xu2017end}       & \cmark & \xmark & 0.1429425 \\
Maqueda et al.  ResNet50 \cite{maqueda2018event}         & \xmark & \cmark & 0.0715585 \\
Yuhuang et al. \cite{hu2020ddd20}  & \cmark & \xmark & 0.0977384 \\
Yuhuang et al. \cite{hu2020ddd20} & \xmark & \cmark & 0.11397   \\
Yuhuang et al. \cite{hu2020ddd20} & \cmark & \cmark & 0.0720821 \\
DRFuser self-attention Resnet 50 & \cmark & \cmark & 0.05192   \\ \bottomrule
\end{tabular}%
}
\end{table}
\begin{table}[]
\centering
\caption{The detailed framework of decoder of the proposed network DRFuser.}
\label{decoder}
\resizebox{10cm}{!}{%
\begin{tabular}{@{}l|c|c@{}}
\toprule
                                                         & Layers             & Input/Output channel size \\ \midrule
\multirow{14}{*}{Decoder with ResNet-50 encoder} & Convolution 2D     & 2048/1024                 \\
                                                         & BatchNormalization & 1024/1024                 \\
                                                         & ReLu               & 1024/1024                 \\
                                                         & Convolution 2D     & 1024/512                  \\
                                                         & BatchNormalization & 512/512                   \\
                                                         & DropOut            & 512/512                   \\
                                                         & ReLu               & 512/512                   \\
                                                         & Convolution 2D     & 512/256                   \\
                                                         & BatchNormalization & 256/256                   \\
                                                         & ReLu               & 256/256                   \\
                                                         & DropOut            & 256/256                   \\
                                                         & Linear layer       & 256*h*w/512               \\
                                                         & ReLu               & 512/512                   \\
                                                         & Linear Layer       & 512/1                     \\ \midrule
\multirow{7}{*}{Decoder with ResNet-34 encoder}  & Convolution 2D     & 512/256                   \\
                                                         & BatchNormalization & 256/256                   \\
                                                         & ReLu               & 256/256                   \\
                                                         & Dropout            & 256/256                   \\
                                                         & Linear layer       & 256*h*w/512               \\
                                                         & ReLu               & 512/512                   \\
                                                         & Linear Layer       & 512/1                     \\ \bottomrule
\end{tabular}%
}
\end{table}

\subsection{Results}
The efficacy in terms of quantitative and qualitative analysis for the proposed method is performed on our collected dataset, DDD dataset, and Carla Eventscape dataset, respectively. RMSE and MAE scores are computed for our quantitative experimental evaluation's proposed and baselines methods. Table-\ref{Table-1} shows the RMSE and MAE scores for our collected dataset. Both RMSE and MAE are negative-oriented scores, which means lower values are better. In Table-\ref{Table-1}, the proposed method is quantitatively compared with the baselines, which includes additive and no-attention models. In our experimentation, we have evaluated the results with the ResNet-34 and ResNet-50 backbone. The experimental results illustrate better results with the ResNet-50 backbone. In addition, the steering angle prediction using the single modality is also performed to provide a sanity check in terms of the fusion of event and frame-based RGB data and illustrate the efficacious of the proposed method in learning end-to-end lateral control. Moreover, the proposed network DRFuser is extensively evaluated on DDD and Carla Eventscape datasets. Table-\ref{Table-2} and Table-\ref{Table-3} shows the quantitative evaluation of DRFuser along with baseline method and single modality evaluation on DDD and Carla Eventscape datasets, respectively. It is to be noted, to the best of our knowledge, only the DDD dataset is utilized in research for the steering angle prediction. Table-\ref{Table-4} shows the quantitative comparison between the proposed DRFuser and state-of-the-art methods. For better evaluation, in Table-\ref{Table-4} event and frame-based RGB data fusion, as well as single modality results, are illustrated. As the focus of this paper is to explore the fusion of event and frame-based RGB data for learning the lateral control in the form of steering angle, the proposed DRFuser achieves a $0.05192$ RMSE score in contrast to $0.0720821$ RMSE score of the state-of-the-art \cite{hu2020ddd20} method. Similarly, the experimental results outperformed the state-of-the-art singular modalities scores. 
% The Table-\ref{Table-1},\ref{Table-2},\ref{Table-3} illustrate the better performance of context-based self-attention as compared to no-attention and additive attention. From the experimental analysis the proposed method has achieved the improvement of $27.33\%$,$25.42\%$ and $39.4\%$ on the collected dataset, DDD, and Carla Eventscape dataset, respectively in contrast to additive attention. This shows that self-attention is much more equipped to manage global context over long than additive attention.
\par 
To further investigate the evaluation of the proposed method DRFuser, a quantitative comparison is performed between the proposed method and baseline as illustrated in Fig.\ref{quantitative-proposed} for our collected dataset, DDD, and Carla Eventscape dataset. The graphs in Fig.\ref{quantitative-proposed} show the prediction results for the proposed method and baseline. The prediction results for the additive-attention baseline over-estimate the prediction result in contrast to the proposed DRFuser method. Similarly, to analyze the effect of fusion in comparison to single modality prediction, a quantitative evaluation is performed, as shown in Fig.\ref{quantitative-comp}. The event camera data, when utilized as a standalone for the steering angle prediction, has not produced promising outcomes as in the case with only frame-based RGB data and DRFuser on all three datasets. The fusion results show better efficacy since the event camera provides complementary information to the frame-based RGB camera, as illustrated in Fig.\ref{quantitative-comp}. It is to be mentioned here that we compared only ResNet-50 based backbone results, as the proposed method has shown better efficacy with the ResNet-50 backbone. We visualized the prediction results for all three datasets with the proposed DRFuser method for the qualitative evaluation, as illustrated in Fig.\ref{visualization}. The Fig.\ref{visualization} only shows a few frames for all three datasets; however, the complete visualization is available at \url{https://github.com/azamshoaib/DRFuser}.

\subsection{Ablation Studies}
To further investigate the efficacy of the proposed method, we expand our analysis by designing the early and late fusion approaches using the same modalities as utilized by the proposed method.
\begin{figure}[t]
      \centering
      \includegraphics[width=12cm]{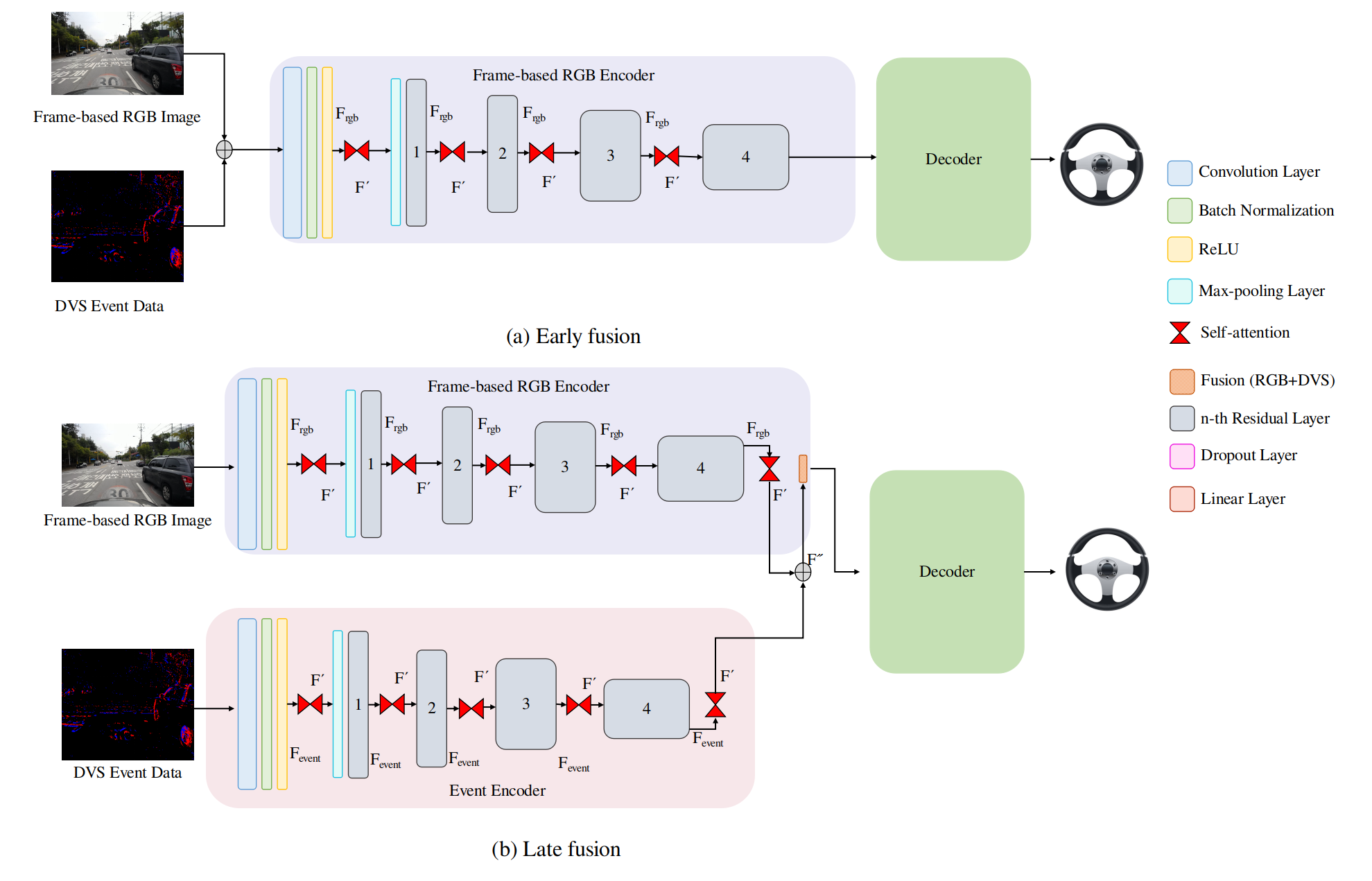}
      \caption{The visualization of prediction of steering angle in comparison of ground-truth, (a) shows collected dataset, (b) shows Davis driving data  and (c) shows Carla EventScape dataset. }
      \label{early-late}
\end{figure}

\begin{figure}[t]
      \centering
      \includegraphics[width=11cm]{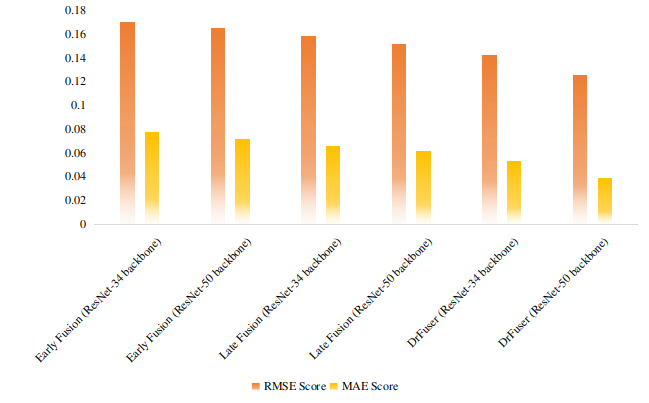}
      \caption{Comparison of early and late fusion approaches with the proposed method on our collected dataset.}
      \label{early-late-our}
\end{figure}
\subsubsection{Comparison of early and late fusion approaches with the proposed method}
\begin{figure}[t]
      \centering
      \includegraphics[width=11cm]{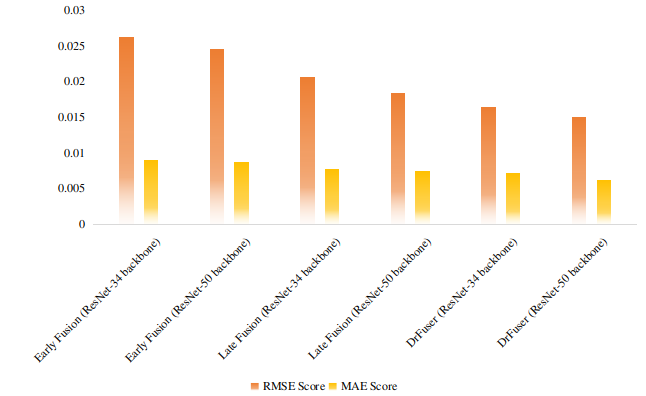}
      \caption{Comparison of early and late fusion approaches with the proposed method on DDD dataset.}
      \label{early-late-ddd}
\end{figure}

\begin{figure}[t]
      \centering
      \includegraphics[width=11cm]{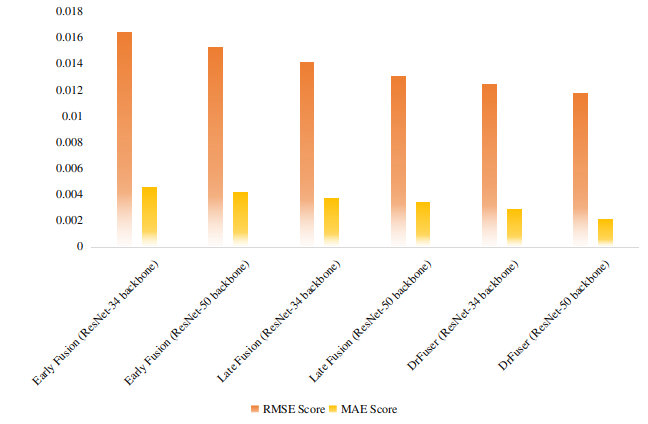}
      \caption{Comparison of early and late fusion approaches with the proposed method on EventScape dataset.}
      \label{early-late-eventscape}
\end{figure}

Multi-modal fusion can be grouped into early, late, and intermediate fusion approaches. In the context of early fusion, the multi-modal features are fused at the input level before the learning algorithm \cite{ramachandram2017deep}. Besides, in the late fusion, the features from the multi-modal are fused at the decision level. However, studies from neuroscience suggest that intermediate fusion could provide the necessary assistance in learning the feature representation from multi-modalities \cite{schroeder2005multisensory} \cite{macaluso2006multisensory}. In order to evaluate the efficacy of the proposed method and analysis, early and late fusion approaches are designed using frame-based RGB and event cameras as modalities. Fig.\ref{early-late}(a) and Fig.\ref{early-late}(b) illustrate the early and late fusion architectures, respectively. The frame-based RGB and event data are fused in the early fusion network by stacking both modalities. In the early fusion, the encoder follows the same architecture as utilized by the proposed network. We have employed the RMSE and MAE as evaluation metrics for the steering angle prediction for the quantitative evaluation of the early fusion with the proposed method. Fig.\ref{early-late-our}, Fig.\ref{early-late-ddd} and Fig.\ref{early-late-eventscape} illustrate the quantitative results of early fusion on our collected,  DDD and EventScape datasets, respectively.

\par 
In the context of the late fusion, the input modalities are fed to the unimodal learning architecture for the feature representation. Later, the features are fused for the final decision of predicting the steering angle. In our experimental analysis, the later fusion performs better in contrast to the early fusion method. The justification of this behavior relies on the architectural structure of CNN that has been carefully designed over the past few years for the unimodal stream. However, the proposed method's efficacy in terms of RMSE and MAE is better than the early and late fusion approaches. Fig.\ref{early-late-our}, Fig.\ref{early-late-ddd} and Fig.\ref{early-late-eventscape} show the quantitative results of late fusion on the three datasets, respectively. It is to be mentioned here that we have employed the same settings for decoder as introduced in the proposed method in both early and late fusion.
\par 
In the experimental analysis of fusing the frame-based RGB and event data using the early fusion approach on our collected dataset, both the backbone network ResNet-34 and ResNet-50 are employed for the fair quantitative evaluation with the proposed method. In the case of early fusion with ResNet-34, RMSE score of $0.1711$ and MAE score of $0.0785$ is achieved in contrast to RMSE and MAE score of $0.1430$ and $0.0540$ respectively for the DRFuser with ResNet-34 backbone. Similarly, the proposed method with ResNet-50 backbone with RMSE and MAE scores of $0.1266$ and $0.0396$ surpasses the early fusion with ResNet-50 backbone having RMSE and MAE scores of $0.1655$ and $0.0724$, respectively. Furthermore, the early fusion results on DDD and EventScape datasets are also quantified with the proposed method using both ResNet-34 and ResNet-50 backbones, as illustrated in  Fig.\ref{early-late-ddd} and Fig.\ref{early-late-eventscape} respectively. For instance, the early fusion with ResNet-34 on the DDD dataset gives the RMSE of $0.02637$ and MAE score of $0.00911$ respectively; in contrast, the DRFuser with ResNet has achieved the RMSE and MAE scores of $0.01654$ and $0.00721$ respectively. Furthermore, the proposed DrFuser with ResNet-50 on the DDD dataset has also surpassed the early fusion with ResNet-50 has the RMSE and MAE scores of $0.02463$ and $0.00887$, respectively, in contrast to DRFuser-ResNet-50 RMSE score of $0.01519$ and MAE score of $0.00631$. In the case of the EventScape dataset, the early fusion with ResNet-34 backbone has achieved the RMSE and MAE scores of $0.0165$ and $0.0046$, respectively. Similarly, with the ResNet-50 backbone, the early fusion gives the RMSE score of $0.0153$ and the MAE score of $0.0042$. These early fusion scores with both backbones on the EventScape dataset have lower RMSE and MAE scores in contrast to the DRFuser with ResNet-34 and ResNet-50 backbone.
\par 
The quantitative results in terms of RMSE and MAE as evaluation metrics for late fusion perform slightly better in contrast to the early fusion approaches rather than the proposed method. For our collected dataset, the late fusion approaches with ResNet-34 and ResNet-50 have achieved the RMSE scores of $0.0142$ and $0.0131$ respectively, in contrast to the DRFuser with ResNet-34 and ResNet-50 RMSE scores of $0.0125$ and $0.0118$. Furthermore, the MAE score of the proposed method with both ResNet-34 and ResNet-50 backbones have better scores in contrast to late fusion approaches. For the DDD dataset, the late fusion approach with ResNet-34 backbone has achieved the RMSE score of $0.02071$ and MAE score of $0.00789$ in contrast to the DRFuser with ResNet-34 backbone having RMSE and MAE scores of $0.01654$ and $0.00721$ respectively. Similarly, for the ResNet-50 backbone, the proposed DRFuser has better RMSE and MAE scores of $0.01519$ and $0.00631$ than the late fusion approach with ResNet-50 having RMSE and MAE scores of $0.01853$ and $0.00754$ respectively. Furthermore, on the EventScape dataset, the late fusion with ResNet-34 and ResNet-50 have achieved the RMSE scores of $0.0142$ and $0.0131$, respectively. The MAE scores of late fusion with both backbones ResNet-34 and ResNet-50 are $0.0038$ and $0.0035$. These RMSE and MAE scores illustrate the lower performance of late fusion compared to the proposed method.

\section{Conclusion}
This paper proposed a novel architecture DRFuser self-attention to fuse frame-based RGB and event data. Here we have shown how deep neural networks benefit from the dynamic response of event cameras and accurately predict vehicle steering angle for a wide range of conditions. Moreover, we have shown the efficacy of the self-attention layer in a deep neural network compared to the element-wise addition of features and using additive attention to fuse the data from two modalities. Moreover, we have conducted an ablation study on the architecture configuration of early, dense and late fusion. The desnse fusion of data is robust in learning data representation and give accurate results. 
Our approach is adapted to process event data as image representation so that a deep neural network can easily be applied. The experimental results show the robustness of DRFuser in comparison to other state-of-the-art methods. 
\par
In future work, we aim to exploit the event data for raining and fog condition and improve the representation of event data to include dense features representation. 

\backmatter

% \bmhead{Supplementary information}

% If your article has accompanying supplementary file/s please state so here. 

% Authors reporting data from electrophoretic gels and blots should supply the full unprocessed scans for key as part of their Supplementary information. This may be requested by the editorial team/s if it is missing.

% Please refer to Journal-level guidance for any specific requirements.

\bmhead{Acknowledgments}

This work was partly supported by Institute of Information \& Communications Technology Planning \& Evaluation (IITP) grant funded by the Korea government (MSIT) (No.2014-3-00077, AI National Strategy Project) and the National Research Foundation of Korea (NRF) grant funded by the Korea government (MSIT) (No. 2019R1A2C2087489), and Ministry of Culture, Sports and Tourism (MCST), and Korea Creative Content Agency (KOCCA) in the Culture Technology (CT) Research \& Development (R2020070004) Program 2022.

\bibliography{sn-article}% common bib file


\end{document}